\documentclass{ifacconf}

\usepackage{graphicx}  
\usepackage{enumitem}
\usepackage{color}
\usepackage{subfigure}
\usepackage{multicol}
\usepackage{nicefrac}
\usepackage{amsmath, amssymb}
\usepackage{listings}
\makeatletter
\let\old@ssect\@ssect 
\makeatother

\usepackage{natbib}
\usepackage[hidelinks]{hyperref}

\makeatletter
\def\@ssect#1#2#3#4#5#6{%
  \NR@gettitle{#6}
  \old@ssect{#1}{#2}{#3}{#4}{#5}{#6}
}
\makeatother

\newcommand{\trnsp}{^\top}

\begin{document}
\begin{frontmatter}

\title{\mbox{\hspace{-22pt} Beyond Occam's Razor in System Identification:} 
\mbox{Double-Descent when Modeling Dynamics \thanksref{footnoteinfo}}} 

\thanks[footnoteinfo]{This research was financially supported by the projects  \emph{Learning flexible models for nonlinear dynamics} (contract number: 2017-03807), \emph{NewLEADS -- New Directions in Learning Dynamical Systems} (contract number: 621-2016-06079), by the Swedish Research Council, by the Brazilian research agency CAPES and by \emph{Kjell och M{\"a}rta Beijer Foundation}.}

\author[First]{Ant\^onio H. Ribeiro}
\author[Second]{Johannes N. Hendriks}
\author[Second]{Adrian G. Wills}
\author[Third]{Thomas B. Sch\"on}

\address[First]{Department of Computer Science, Federal University of Minas Gerais, 31270-901 Belo Horizonte, Brazil (antoniohorta@dcc.ufmg.br)}
\address[Second]{School of Engineering, The University of Newcastle, Callaghan NSW, Australia (\{johannes.hendriks,adrian.wills\}@newcastle.edu.au)}
\address[Third]{Department of Information Technology, Uppsala University, Uppsala, Sweden (thomas.schon@it.uu.se)}

\begin{abstract}
System identification aims to build models of dynamical systems from data. Traditionally, choosing the model requires the designer to balance between two goals of conflicting nature; the model must be rich enough to capture the system dynamics, but not so flexible that it learns spurious random effects from the dataset. It is typically observed that the model validation performance follows a U-shaped curve as the model complexity increases. Recent developments in machine learning and statistics, however, have observed situations where a ``double-descent'' curve subsumes this U-shaped model-performance curve. With a second decrease in performance occurring beyond the point where the model has reached the capacity of interpolating---i.e., (near) perfectly fitting---the training data. To the best of our knowledge, such phenomena have not been studied within the context of dynamic systems. The present paper aims to answer the question: ``Can such a phenomenon also be observed when estimating parameters of dynamic systems?'' We show that the answer is yes, verifying such behavior experimentally both for artificially generated and real-world datasets.
\end{abstract}


\end{frontmatter}

\section{Introduction}
\begin{figure}[b] 
\vspace{-5pt}
\noindent \footnotesize{\textbf{This preprint will appear in the Proceedings of the 19th IFAC Symposium in System Identification. Please cite:}}
\begin{lstlisting}
@inproceedings{ribeiro_occam_2021,
author={Ant\^onio H. Ribeiro and Johannes N. Hendriks and Adrian G. Wills and Thomas B. Sch\"on},
title={{B}eyond {O}ccam's {R}azor in {S}ystem {I}dentification: {D}ouble-{D}escent when {M}odeling {D}ynamics}, 
year={2021},
booktitle={{P}roceedings of the 19th {IFAC} {S}ymposium in {S}ystem {I}dentification ({SYSID})}
}
\end{lstlisting}
\vspace{-5pt}
\end{figure}

\begin{figure}
    \centering
    \subfigure[U-shape performance]{\includegraphics[width=0.49\linewidth]{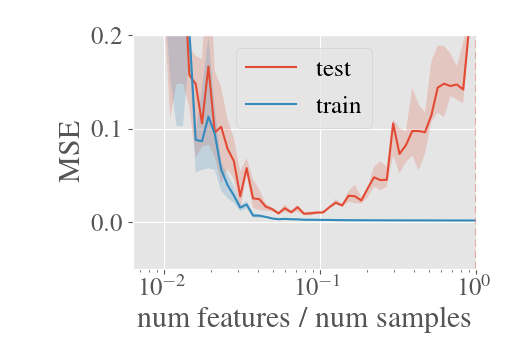}} 
    \subfigure[double-descent performance]{\includegraphics[width=0.49\linewidth]{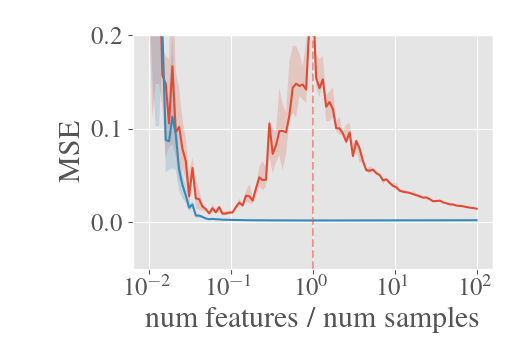}}
    \caption{\textbf{Double-descent in the CE8 benchmark.} We show the one-step-ahead prediction error for a nonlinear ARX model in the CE8 benchmark~\citep{wigren2017coupled}.  In (a), we show the \textit{``classical''} regime and a U-shape performance curve. In (b), the vertical dashed line indicates the transition from the \textit{``classical''} regime to the interpolation regime: when the model capacity is large enough to perfectly fit the data (i.e. zero training error). While the U-shape model performance is observed in the \textit{``classical''} regime and the error is large at the interpolation threshold, increasing the capacity beyond this point leads to decreasing the error in the test set. The experiment is detailed in Section~\ref{sec:additional-examples}.}
    \label{fig:double_descent_illustration}
\end{figure}

Traditionally, there is a trade-off when choosing the model complexity: the model must be rich enough to capture the dynamics in the data, but not so flexible that it learns spurious random effects. This corresponds to the classical U-shape performance curve that is typically considered when choosing model complexity (see Fig.~\ref{fig:double_descent_illustration}(a)).

As we increase the model flexibility, it is possible to reach a point where the training error is zero. At this point, the model achieves low bias and large variance. That is, the model does not generalise well and will perform poorly on an unseen test dataset. However, recent work by~\citet{belkin_reconciling_2019} has shown that if we continue to increase the model complexity beyond this point, the model can eventually start to generalise well again. That is, the bias remains low and the variance starts to decrease again. Figure~\ref{fig:double_descent_illustration}(b) shows this behavior. This is known as the \textit{double-descent} curve and it subsumes the traditional U-shaped curve.

This seemingly counterintuitive behavior has been explored in related fields. In this paper, we seek to demonstrate it on data from dynamic systems using Nonlinear ARX (Auto-Regressive with eXogenous inputs) models. We give examples of different models and regularization strategies that can lead to its observation.

\vspace{-10pt}
\subsubsection{\textbf{Related work and historical development}}

Deep neural networks have achieved state-of-the-art solutions for many tasks~\citep{lecun_deep_2015}. These models, however, often have millions or, even, billions of parameters \citep{tan_efficientnet_2019}, which seems at odds with basic system identification (and statistics) tenets and the parsimonious principle:  \textit{``model structure should give enough flexibility to model the system dynamics but not more''}. Furthermore, while the number of parameters is not always a perfect measure of the capacity of the model, deep learning models have been shown to have enough capacity to fit a training set labeled at random~\citep{zhang_understanding_2017}.  It has also been observed that these models seem to indefinitely display increased performance as the model size increases~\citep{tan_efficientnet_2019}.

\citet{belkin_reconciling_2019} reconciled this phenomenon with the more traditional bias-variance trade-off paradigm. There, model generalization is studied in the interpolation regime, i.e., for which the model has enough capacity to perfectly (or almost perfectly) fit the training data. Although the learned predictors obtained at the interpolation threshold typically have high risk, increasing the capacity beyond this point leads to decreasing risk, sometimes achieving better performance than in the ``classical'' regime.

The double-descent performance curve has been experimentally observed in diverse machine learning settings: \citet{belkin_reconciling_2019} show it for random Fourier features, random forest and shallow networks, while~\citet{nakkiran_deep_2020} show the same phenomenon for transformers and convolutional network models. In a different line of work, the phenomenon has been studied theoretically. \citet{hastie_surprises_2019, mei_generalization_2019} provide asymptotic guarantees for regression with random features and \citet{bartlett_benign_2020} provides finite sample generalization bounds. 

\vspace{-10pt}
\subsubsection{\textbf{Contributions}}
Here, we study double-descent in the usual system identification setting where the data comes from the input and output of a dynamical system. We provide experimental evidence it can indeed be observed in this scenario (cf. Fig.~\ref{fig:double_descent_illustration}) with experiments on both artificially generated and real-world data sets. We also discuss the mechanisms that can yield the observation of the phenomenon.

\vspace{-10pt}
\subsubsection{\textbf{Code availability}} The code for reproducing the experiments is available at:\\
{\href{https://github.com/antonior92/narx-double-descent}{\hspace{13pt}https://github.com/antonior92/narx-double-descent}}

\vspace{-6pt}

\section{Motivation example}
\label{sec:motivation-examples}
\begin{figure*}[t!]
    \centering
    \subfigure[one-step-ahead MSE]{\includegraphics[width=0.32\linewidth]{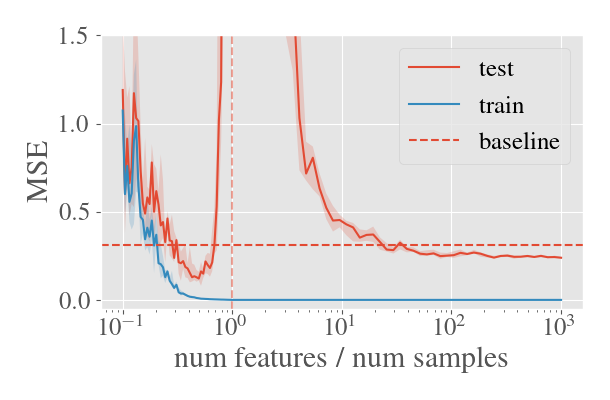}}
    \subfigure[free-run simulation MSE]{\includegraphics[width=0.32\linewidth]{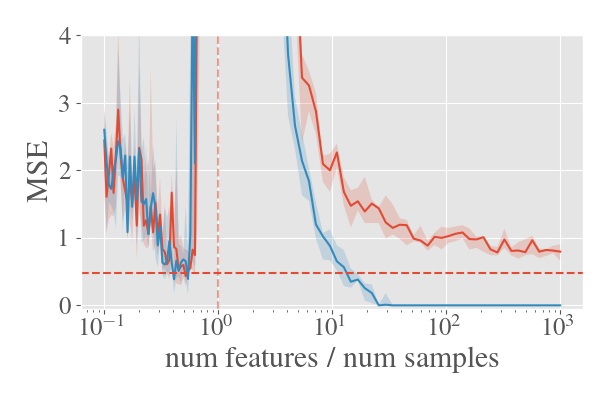}}
    \subfigure[parameter norm]{\includegraphics[width=0.32\linewidth]{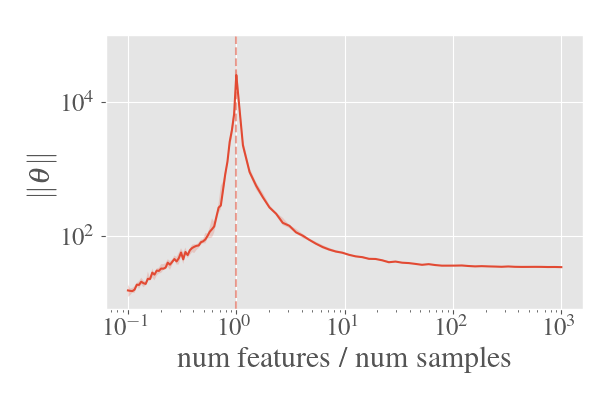}}
    \label{fig:random_features_dd}
    \caption{\textbf{Double-descent performance curve when modeling the nonlinear system \eqref{eq:chen_model}.} We display the performance of the minimum-norm least-square solution for RFF models ($\gamma = 0.6$) on data generated from~\eqref{eq:chen_model}, with $\sigma_v=0.1$ and $\omega_c = 0.7$. In (a) and (b) we show the train and test MSE. In (c), we show the parameter norm of the corresponding solutions. We keep the number of samples in the training and test datasets constant ($T =400$ and $T' = 100$) and vary the number of parameters $m$ in a log-uniform grid in $[10^{-1}T, 10^3T]$, repeating the same experiment 10 times in each of the 100 points of the grid. The solid line is the median of the 10 experiments and the shaded region delimits the inter-quartile range (i.e the range between $25\%$ and $75\%$ percentiles). The dashed horizontal line gives the test performance of a linear ARX model (with the same delays) that is used as baseline.}
\end{figure*}

We start by presenting a simple example for which the double-descent performance curve can be observed.

\subsubsection{\textbf{Dataset}}
Consider the nonlinear system presented by \citet{chen_non-linear_1990}. Let $u_t\in \mathbb{R}$ and $y_t\in \mathbb{R}$ denote the input and output, respectively. The system output is given by the  difference equation
\begin{equation}\label{eq:chen_model}
\begin{split}
    y_t =& \left(0.8 - 0.5e^{-y^{2}_{t-1}}\right)y_{t-1} - \left(0.3+0.9e^{-y^{2}_{t-1}}\right)y_{t-2}\\ &+ u_{t-1}+0.2u_{t-2}+0.1u_{t-1}u_{t-2}+v_t,
\end{split}
\end{equation}
where $v_t \sim \mathcal{N}(0,\sigma_v^2)$ represents the process noise and $u_t$ is generated by  applying a low-pass filter with cutoff frequency $\omega_c$ to a Gaussian white noise signal with unitary variance. We generate $T$ samples for training the model and a hold-out test set of $ T'$ samples to evaluate its performance on unseen data.

\subsubsection{\textbf{Model}} We use a nonlinear ARX model~\citep{ljung_system_1998} to identify the proposed system. Let us denote
\begin{equation}
    \label{eq:specific_x}
   x_t = (u_{t-1}, u_{t-2}, y_{t-1}, y_{t-2}).
\end{equation}
We consider a linear-in-the-parameter model for predicting the output from the observed past input/output values
\begin{equation}
    \label{eq:nonlin-map}
    \hat{y}_t = f(x_t) = \sum_{i=1}^m \theta_i \phi_i(x_t). 
\end{equation}

Given the training sequence $\{(u_t, y_t), t=1, \cdots, T\}$, the model is estimated by finding the values $\theta_i$ that minimize
\begin{equation}
    \label{eq:narx-estimation}
    \frac{1}{T}\sum_{t = 1}^T\left\| y_t - \sum_{i=1}^m \theta_i \phi_i(x_t)\right\|^2.
\end{equation}
Or, equivalently, in matrix form, by finding the vector $\theta \in \mathbb{R}^m$ that minimizes
\begin{equation}
     \label{eq:narx-estimation-matrix}
    \frac{1}{T}\|y - \Phi \theta\|^2,  
\end{equation}
where $\Phi \in\mathbb{R}^{T\times m}$ is the matrix containing  $\phi_i(x_t)$ at position $(t, i)$ and $y \in \mathbb{R}^T$ is the vector of outputs. Indeed, finding the optimal parameter here is an ordinary least-squares problem and its analytical solution is
\begin{equation}
\label{eq:ls-sol}
\hat{\theta} = (\Phi\trnsp \Phi)^{+}\Phi\trnsp y,
\end{equation}
where $(\Phi\trnsp \Phi)^{+}$ denotes the Moore-Penrose pseudo-inverse of $\Phi\trnsp \Phi$.  Next, we detail the choice of the nonlinear feature map used in this example.

\subsubsection{\textbf{Random Fourier features} {\bf (RFF)}} \hspace{-5pt} We use the feature map introduced by 
\citet{rahimi_random_2008} to approximate the reproducing kernel Hilbert space (RKHS) defined by the Gaussian kernel ${K(x, x') = \exp(-\gamma \|x - x'\|^2)}$. More precisely, the features are generated as
\begin{equation}
    \phi_i(x) = \sqrt{\frac{2}{m}}\cos(w_i\trnsp x + b_i),
\end{equation}
where $w_i\in\mathbb{R}^n$ is a vector with each element sampled independently from $\mathcal{N}(0, 2\gamma)$ and $b_i\in\mathbb{R}$ is sampled from a uniform distribution $\mathcal{U}[0, 2\pi)$. Here,~$\gamma$ is a tunable hyper-parameter of the method.

\subsubsection{\textbf{Metrics}}
We use the \textit{mean squared error (MSE)} as performance metric,
\begin{equation}
{\rm MSE} = \frac{1}{T}\sum_{t = 1}^T\|\hat{y}_{t} - y_t\|^2.
\end{equation}
We refer to the \textbf{one-step-ahead MSE} when the one-step-ahead prediction $\hat{y}_{t}$ is used in the computation. By one-step-ahead we refer to predictions computed as in Eq.~\eqref{eq:nonlin-map}, with the observed past inputs being used to predict the current output. On the other hand, we refer to \textbf{free-run-simulation MSE} when the MSE is computed for the simulations $\hat{y}_{t}^{\text{free}}$, obtained by free-run simulating the model. That is, $\hat{y}_{t}^{\text{free}}$ is computed by the recursive formula:
\begin{equation*}
    \hat{y}_{t}^{\rm free} = 
    \begin{cases}
    y_{t}\text{ for }t = 1, \cdots, n_y,\\
     f(u_{t}, \cdots, u_{t-n_u+1}, \hat{y}_{t-1}^{\rm free}, \cdots, \hat{y}_{t-n_y}^{\rm free})\text{ for }t> n_y,
    \end{cases}
\end{equation*}
where the previously predicted outputs (rather than the observed ones) are used to compute the next step.

\subsubsection{\textbf{Results}}
In Fig.~\ref{fig:random_features_dd}, we show the performance of the RFF models on the training and test datasets as a function of the proportion $m/T$, i.e. the number of parameters (features) divided by the total training sequence length. In Fig.~\ref{fig:random_features_dd}(a), we show the one-step-ahead MSE, which displays the U-shape test performance curve followed by a second descent in the test performance, which is the result of performance improvements as we increase the number of features after the interpolation threshold ($m/T = 1$). In Fig.~\ref{fig:random_features_dd}(b), we show the free-run simulation MSE and demonstrate that we can still observe the second descent in test performance for this scenario. It is interesting to note that while the model reaches one-step-ahead training error close to zero for $m/T = 1$---due to the numerical approximation errors\footnote{For $m/T = 1$---the matrix $\Phi$ has a large condition number which yields numerical errors when estimating the parameters. In this example,  $\nicefrac{\sigma_{\max}}{\sigma_{\text{min}}} > 10^6$, where $\sigma_{\max}$ the maximum singular value of $\Phi$. The error is then accumulated through the recurrence in the free-run simulation of the estimated system.}, the free-run simulation MSE on the training data approaches zero only for larger values of $m/T$. Fig.~\ref{fig:random_features_dd}(c) displays the parameter norm $\|\theta\|_2$ as a function of the proportion $m/T$, showing that it peaks at the interpolation threshold and then monotonically decrease. This is something that we will explore later. With this, we finish the presentation of our initial example. Next, we will further explore different mechanisms and settings that give rise to double-descent performance curves.

\section{Linear-in-the-parameters models}
\label{sec:linear-in-the-parameters}

In this section, we further investigate the phenomena in the linear-in-the-parameters setting. We first study options for selecting one solution (over many) in the overparametrized regime and then present additional examples with different datasets and features.

\subsection{Selecting the solution in the interpolation regime}
\label{sec:selecting-the-solution}

In the case where the number of features is larger than the number of measurements, i.e. $m>T$, there are multiple possible solutions for~\eqref{eq:narx-estimation-matrix}. We discuss different choices next.

\subsubsection{\textbf{The minimum-norm solution}} One natural option is to, in the overparametrized case, use the minimum $\ell_2$-norm solution to the problem. That is, the solution
\begin{equation}
\label{eq:min-norm-solution}
\hat{\theta} = \text{arg}\min_\theta \|\theta\|_2 \quad \text{subject to}\quad\Phi\theta = y.
\end{equation}
In the motivation example, we used $\hat{\theta} = (\Phi\trnsp \Phi)^{+}\Phi\trnsp y$, which is equivalent to~\eqref{eq:min-norm-solution} in the overparametrized case (i.e. $m>T$) due to the use of the Moore-Penrose pseudo-inverse, which yields the minimum norm solution by definition.

Fig.~\ref{fig:random_features_dd}(c) displays the parameter norm $\|\theta\|_2$ as a function of the proportion $m/T$, showing that it peaks at the interpolation threshold and monotonically decreases after it. The intuition behind this behavior is that at the interpolation threshold there is a unique solution and usually this solution has a large norm, but as we increase the problem dimension ($m$), the space of possible solutions increases and it becomes possible to find solutions with smaller norm. Thus, an interpretation of the second descent in the performance curve is that increasing the number of parameters yields solutions with smaller parameter norm (cf. Fig~\ref{fig:random_features_dd}), which is a type of inductive bias resulting in a decreasing variance error after the interpolation threshold. This argument was presented, for instance, by~\citet{belkin_reconciling_2019}.

\vspace{-5pt}

\subsubsection{\textbf{The effect of regularization}} 
\begin{figure}
    \vspace{-10pt}
    \centering
    \includegraphics[width=0.78\linewidth]{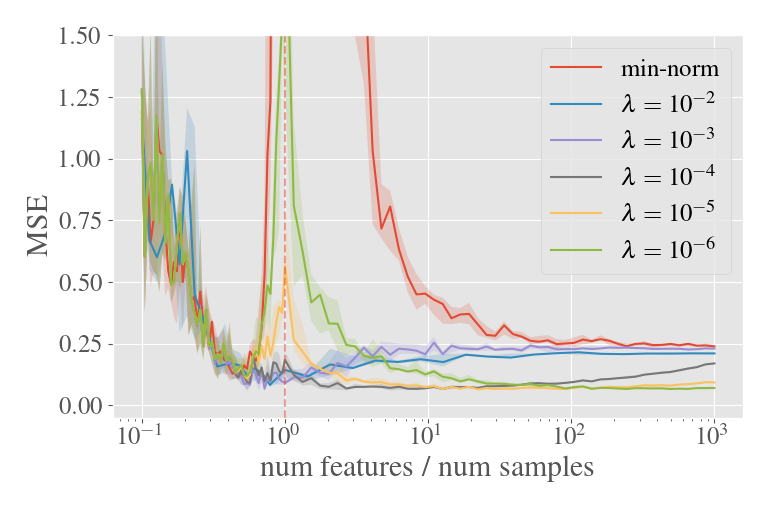}
    \caption{\textbf{Ridge regression with vanishing values of $\lambda$.} The figure displays the one-step-ahead MSE in the test set for RFF models in the nonlinear system~\eqref{eq:chen_model}. The performance curve  labeled ``min-norm'' is obtained by solving problem~\eqref{eq:min-norm-solution} and all the other curves correspond to the solution given in~\eqref{eq:ridge_regression} with progressively smaller values of $\lambda$. The experimental setting is the same as that used for  Fig.~\ref{fig:random_features_dd}. }
    \label{fig:ridge}
\end{figure}

It is common to add a regularization term to the least-square problem~\eqref{eq:narx-estimation-matrix}, penalizing the $\ell_2$ parameter norm. This results in the so-called ridge regression problem, which has a unique solution (even in the interpolation regime) given by
\begin{equation}
     \label{eq:ridge_regression}
    \hat{\theta}_\lambda = \text{arg}\min_\theta\left(\frac{1}{T}\|y - \Phi \theta\|^2  + \lambda \|\theta\|^2\right).
\end{equation}
Fig.~\ref{fig:ridge} displays the performance curve as a function of $m/T$ for different values of $\lambda$. It illustrates that the double-descent becomes more explicit for smaller values of $\lambda$. Indeed, with  $\hat{\theta}_\text{min-norm}$ defined as in~\eqref{eq:min-norm-solution}, it is possible to prove that $\lim_{\lambda\rightarrow 0^+} \hat{\theta}_\lambda = \hat{\theta}_\text{min-norm}$, see~\citet{hastie_surprises_2019}.

\subsubsection{\textbf{Ensembles}}
\begin{figure}
    \vspace{-10pt}
    \centering
    \subfigure[one-step-ahead MSE]{\includegraphics[width=0.49\linewidth]{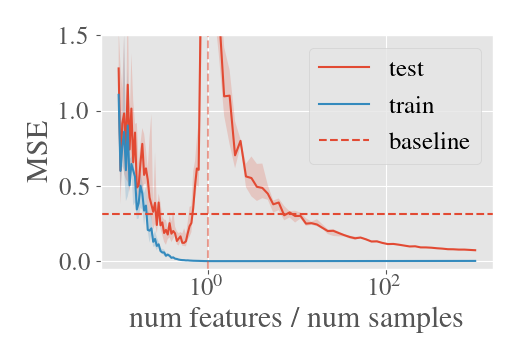}}
    \subfigure[free-run simulation MSE]{\includegraphics[width=0.49\linewidth]{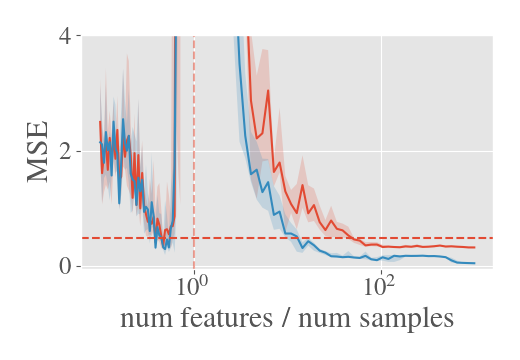}}
    \caption{\textbf{Ensembles after the interpolation threshold}. The figure displays the train and test MSE. After the interpolation threshold we use Eq.~\eqref{eq:ensemble-linear-in-param} for an ensemble with $B=1000$ different solutions. The experimental settings are the same as those for Fig~\ref{fig:random_features_dd}. }
    \label{fig:random_features_dd_ensembles}
\end{figure}

Another mechanism for choosing solutions beyond the interpolation threshold that also yields increased performance as the model class is enlarged is the use of ensembles. Here we give one example of an ensemble that is linear-in-the-parameters and in Section~\ref{sec:ensembles-and-rf} we give an example that is not.

For $m>T$, assume that we select a subset of indices $\mathcal{S}_{b}\subset \{1, \cdots, m\}$, where the cardinality of this set is $|\mathcal{S}_{b}| = T$. Let ${\rm S}_b\in \mathbb{R}^{m \times T}$  be the selection matrix obtained by selecting the columns of the identity matrix $I_m$ corresponding to the indices in $\mathcal{S}_{b}$. The matrix $\Phi{\rm S}_b$ is a square matrix in $\mathbb{R}^{T \times T}$ and we can (uniquely) find the parameters $\hat{\theta}_b \in \mathbb{R}^T$ by solving the linear system $(\Phi{\rm S}_b) \hat{\theta}_b = y$. In this case, ${\rm S}_b\trnsp\hat{\theta}_b$ is one solution of the overparametrized least-square problem defined in~\eqref{eq:narx-estimation-matrix}. Assume that we repeat the same procedure $B$ times for different selection matrices and get the average solution
\begin{equation}
\label{eq:ensemble-linear-in-param}
    \hat{\theta}^{\rm ens} = \frac{1}{B}\sum_{b=1}^B {\rm S}_b\trnsp\hat{\theta}_b.
\end{equation}
It is easy to verify that this is still a solution to~\eqref{eq:narx-estimation-matrix}. In Fig.~\ref{fig:random_features_dd_ensembles} we show that this procedure is an alternative mechanism that also yields a second descent in performance after the interpolation point. For numerical stability, rather than solving the linear system $(\Phi{\rm S}_b) \hat{\theta}_b = y$ we solve a ridge regression problem with very small value of $\lambda$.  For instance, in Fig.~\ref{fig:random_features_dd_ensembles}, we use $\lambda = 10^{-7}$.

Our focus here is just to present ensembles as an alternative mechanism for observing the double-descent performance curve after the interpolation threshold. Hence, in Fig.~\ref{fig:random_features_dd_ensembles} ensembles are used only after the interpolation point. Nonetheless, we would like to highlight that ensemble models can boost the performance even before the interpolation threshold. We refer the reader to  \citet{lejeune_implicit_2020} for an in-depth analysis of ensembles of ordinary least squares  in the underparametrized regime.

\subsection{Additional examples}
\label{sec:additional-examples}
\begin{figure}
    \centering
    \vspace{-10pt}
    \includegraphics[width=0.78\linewidth]{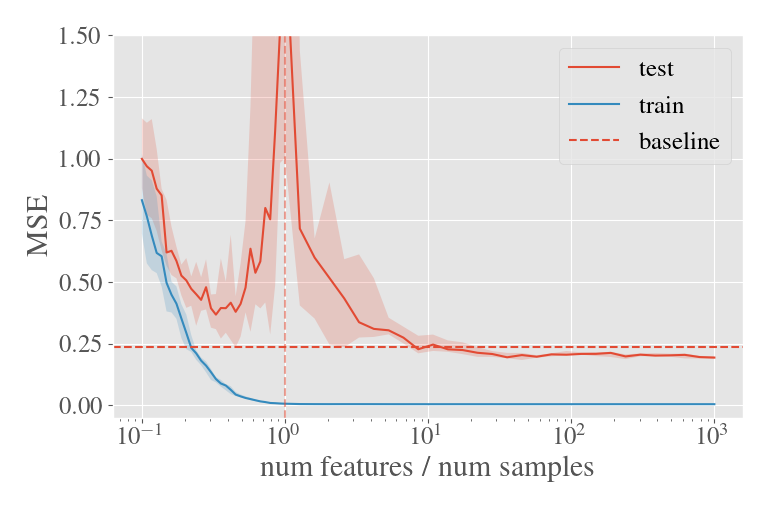}
    \caption{\textbf{RBF networks.} The one-step-ahead MSE in the training and test set for RBF network models ($\gamma = 0.25$ and $\eta=5$) in the nonlinear system from \citet{chen_non-linear_1990}. The data was generated using Eq.~\eqref{eq:chen_model}, with $\sigma_v=0.1$, $T =400$ and $T' = 100$. After the interpolation threshold, we use the ensemble strategy for choosing the solution (as in Fig.~\ref{fig:random_features_dd_ensembles}), with $B=2000$ and $\lambda = 10^{-14}$. }
    \label{fig:random_features_dd_rbfnet}
\end{figure}

Here we give alternative linear-in-the-parameter scenarios where we have experimentally observed a  double-descent performance curve. We have considered RFF in all the examples so far. Next, we present an alternative definition of nonlinear feature maps which result in similar behavior.

\subsubsection{\textbf{Radial basis function {\bf (RBF)} network}} For RBF networks, given the centers $c_i\in\mathbb{R}^n$, the features are generated as
\begin{equation}
    \phi_i(x) = \exp(-\gamma \|x - c_i\|).
\end{equation}
This class of functions are universal approximators in a compact subset~\citep{park_universal_1991}, and can be formulated as non-convex optimization problems when $c_i$ are treated as free optimization parameters. Here, however, we choose the centers at random, sampling them from $\mathcal{N}(0, \eta I_n)$. This yields a hypothesis class that fits in Eq.~\eqref{eq:nonlin-map} and can be solved using ordinary least squares. In Fig.~\ref{fig:random_features_dd_rbfnet}, we provide the performance curve of RBF networks when modeling the nonlinear system described in~\eqref{eq:chen_model}.

We also tested the phenomena using a real world dataset, described next.

\subsubsection{\textbf{Coupled electrical (CE8) drives benchmark}}

We also observe the phenomena in the dataset collected from the operation of coupled electric drives~\citep{wigren2017coupled}. The system consists of two electric motors that drive a pulley using a flexible belt. The pulley is held by a spring and the angular speed is measured by a pulse counter. The system to be identified takes as input the control signal sent to both motors (which are the same) and should predict as output the angular speed. The pulse counter is insensitive to the sign of the angular velocity, which creates an ambiguity in the measurements and makes the problem harder.

We use two sequences of 10 seconds to develop the model. The sequences were collected from the above system operating with inputs uniformly distributed in amplitude. The first 60\% of the measurements are used for training and the remaining 40\%, for testing. In Fig.~\ref{fig:double_descent_illustration}, we display the training and test MSE for RFF models ($\gamma = 0.2$). After the interpolation threshold, we use the ensemble strategy, with $B=2000$ and $\lambda = 10^{-14}$, for choosing the solution.

\section{General nonlinear models}

Let us now consider nonlinear ARX models that cannot be formulated as linear-in-the-parameters problems. As in Section~\ref{sec:motivation-examples}, let $x_t$ be defined as a concatenation of past input and outputs. Given a training set, the problem can be formulated as choosing the function $f$ in the hypothesis  class $\mathcal{F}$ that (exactly or approximately) minimizes
\begin{equation}
    \label{eq:narx-estimation-nonlinear-in-param}
    V = \frac{1}{T}\sum_{t = 1}^T\|f(x_t) - y_t\|^2.
\end{equation}
Assume that $\omega \in \mathbb{R}_+$ is a parameter that controls 
the size of the hypothesis class---i.e. $\mathcal{F}_{\omega_1} \subset \mathcal{F}_{\omega_2}$ if $\omega_1 < \omega_2$. Let $\omega_t$ denote the threshold after which it is possible to select $f$ that yields zero training error (i.e., the model perfectly fits the training data). We refer to the double-descent phenomenon as the situation for which the test error decreases with $\omega$ in the  interpolation regime $\omega\in (\omega_t, \infty)$.

\subsection{Random forests}
\label{sec:ensembles-and-rf}

\begin{figure}
    \centering
    \vspace{-10pt}
    \subfigure[one-step-ahead MSE]{\includegraphics[width=0.49\linewidth]{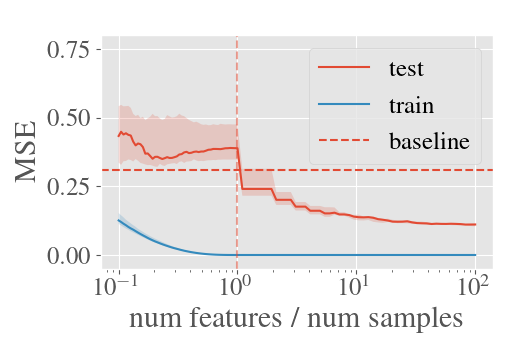}}
    \subfigure[free-run simulation MSE]{\includegraphics[width=0.49\linewidth]{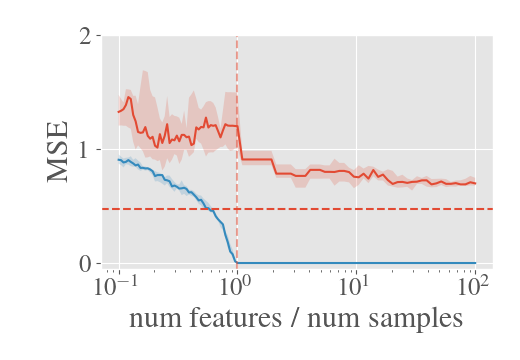}}
    \caption{\textbf{Double-descent for Random Forest models.} The data was generated as in Eq.~\eqref{eq:chen_model}, with $\sigma_v=0.1$. In (a) and (b) we show the train and test MSE. We keep the number of samples of the training and test datasets constant ($T =3000$ and $T' = 100$) and vary the number of leafs $m$ in a log-uniform grid in the interval $[10^{-1}T, 10^{2}T]$.}
    \label{fig:random_forest_dd}
\end{figure}

We can generalize the idea of an ensemble (from Section~\ref{sec:selecting-the-solution}) to this more general scenario. Before the interpolation threshold, the function $f$  which minimizes \eqref{eq:narx-estimation-nonlinear-in-param} is selected from the hypothesis class  $\mathcal{F}$. When $\omega > \omega_t$ there might be multiple possible solutions, so  we pick $B$ different solutions $f_b$ from the hypothesis class, and use the average of their predictions,
\begin{equation}
    \label{eq:bagging}
    f(x) = \frac{1}{B} \sum_{b = 1}^B f_b(x).
\end{equation}
The random forest~\citep{breiman_random_2001} is a popular ensemble method which we will study here. For this model class, $f_b$ is a decision tree. That is, a rooted tree structure is associated with $f_b$ and the output is computed by traveling from the tree root node to one of the leaves (which is associated with a given output). At each node of the tree, a given $x_i$ is used as a decision variable to decide which child node to navigate to. The tree structure, the decision variables and the decision stumps are obtained (i.e., the model is trained) by approximately minimizing \eqref{eq:narx-estimation-nonlinear-in-param} using a greedy algorithm.

The number of leaves of a decision tree provides a natural way to parameterize the capacity of the model. A tree with $m$ leaves corresponds to a piece-wise function consisting of $m$ constant functions and, as such, it can interpolate $m$ data points. To increase the capacity of the model beyond the interpolating threshold, ensembles (averages) of multiple decision trees are used, i.e. Eq.~\eqref{eq:bagging}. Here, the different models $f_b$ are obtained by presenting the data in a different order compared to the (suboptimal) greedy optimization algorithm. We do not use bootstrap resampling (as it is traditionally done). This way, each $f_b$ perfectly fits the training dataset after the interpolation threshold, in agreement with the setup we are interested in.

\subsubsection{\textbf{Results}} In Fig.~\ref{fig:random_forest_dd}, we show the performance as a function of the proportion between the total number of leaves in the random forest and the number of training samples. Showing that increasing the model capacity beyond the interpolation threshold yields continuous improvements in the performance.

\subsection{A note on neural networks}
\label{sec:nn}

In the introduction, we described how the success of deep neural networks was an important reason for digging deeper into the properties of overparametrized models. In this section, we briefly describe some connections between the examples and ideas we presented in this paper and the study of neural networks. We appeal to a recent line of work by~\citet{jacot_neural_2018, chizat_lazy_2019}, which derive approximate models for neural networks where analytical solutions are easier to study. 

Deep neural networks can be understood as black-box parametrized functions $f_\theta(\cdot)$ that are nonlinear in the parameters and for which~\eqref{eq:narx-estimation-nonlinear-in-param} yields a non-convex problem. Let $\theta \in \mathbb{R}^m$ be the neural network parameters, following \citet{chizat_lazy_2019},  we assume that the number of parameters is very large and that training the neural network moves each of them just by a small amount w.r.t. its initialization $\theta_0$. It thus makes sense to linearize the model around $\theta_0$, which yields 
\begin{equation}
f(x; \theta) \approx  f(x; \theta_0) + \nabla f(x; \theta_0)\trnsp\tilde{\theta},
\end{equation}
where $\tilde\theta = \theta - \theta_0$. Hence, denoting $\phi_i = \nabla f(x; \theta_0)$ and assuming  $f(x; \theta_0) \approx 0$ we return to the setup of Section~\ref{sec:linear-in-the-parameters}. 

Another interesting connection, that will be explored next, is the relation between the solution of the gradient descent algorithm and the minimum-norm solution of least-squares problems. Gradient descent algorithm and its variations are popular choices for estimating deep neural network parameters. In the vanilla version, this algorithm iteratively refines the parameter $\theta$ by moving in the opposite direction of the gradient of the cost function $V$,
\begin{equation}
    \label{eq:gradient-descent}
    \theta^{i+1} = \theta^i - \gamma \nabla_{\theta} V (\theta^i),
\end{equation}
where $\gamma$ is the \textit{learning rate} that controls the optimization and $\nabla_{\theta} V (\theta^i)$ denotes the gradient of $V$ evaluated at $\theta^i$ and $\theta^i$ denotes the parameter estimate at the $i^{\text{th}}$ iteration.

The use of the minimum-solution norm solution yields, in the setting from Section~\ref{sec:linear-in-the-parameters}, a second descent in the performance curve in the overparametrized region. The next theorem establishes that if we initialize $\theta^0$ in the row space of $\Phi$, then gradient descent finds the minimum-norm solution of least-squares problems. This is another interesting connection between the setup studied here and the standard deep neural network setup.

\begin{thm}
Let $V(\theta) = \frac{1}{2}\|\Phi \theta + y\|^2$, for $\Phi \in \mathbb{R}^{T \times m}$ a matrix with full row rank. Let $\theta^i$ be the $i^{\text{th}}$ step of the gradient-descent algorithm (defined in Eq.~\eqref{eq:gradient-descent}), initialized with $\theta^0$ in the row space of $\Phi$.  Then, if $\hat{\theta}$ denotes the minimum-norm minimizer of $V(\theta)$ there exist a $\tilde{\gamma}>0$ such that for all $\gamma\in(0, \tilde{\gamma})$, $\theta^i \rightarrow \hat{\theta}$ as $i \rightarrow \infty$.
\end{thm}

\begin{pf}
The above result follows from analytically computing the gradient, plugging it into Eq.~\eqref{eq:gradient-descent}, using the SVD decomposition of $\Phi$, and then taking the limit. \citet[Proposition 1]{hastie_surprises_2019} and 
\citet{de_azevedo_does_2020} provide complete proofs of the statement.
\end{pf} 

While the connections presented here are not exact, they give some insight into the generalization properties of neural network models. Indeed, double-descent performance curves have been experimentally observed in~\citet{belkin_reconciling_2019} for shallow neural networks and in~\citet{nakkiran_deep_2020} for transformers and convolutional neural networks.

\section{Conclusion and Future Work}

In this paper, we have presented the double-descent phenomenon in a system identification framework, giving experimental evidence that it holds for nonlinear ARX models. We also discuss the mechanisms that lead to it.

It is well-known within the system identification community that the assumptions needed to guarantee that the nonlinear ARX estimates are consistent are rather strict, only white process noise can be present. Studying double-descent for ARMAX, output error and other types of models that can handle more general noise types~\citep{ljung_system_1998} is a natural and interesting future direction. Furthermore, the assumption of independent regressors used in \citep{hastie_surprises_2019, mei_generalization_2019, bartlett_benign_2020} and other theoretical analysis of the double-descent curve do not hold in the setup we presented here. Hence, extending the available theoretical results to the presented setup is also another interesting direction.


\end{document}